\documentclass[10pt,twocolumn,letterpaper]{article}

\usepackage[pagenumbers]{cvpr}      

\usepackage{graphicx}
\usepackage{amsmath}
\usepackage{amssymb}
\usepackage{booktabs}
\usepackage{multicol, multirow}
\usepackage{subcaption}
\usepackage{dsfont}
\usepackage{xcolor,colortbl}
\usepackage{tabularx,array,makecell}
\newcolumntype{L}[1]{>{\raggedright\arraybackslash}p{#1}}
\newcolumntype{C}[1]{>{\centering\arraybackslash}p{#1}}
\newcolumntype{Y}{>{\centering\arraybackslash}X} 

\newcommand{\up}[1]{\textcolor{green!60!black}{\ensuremath{\uparrow}#1}}

\definecolor{dioBlueFill}{HTML}{DAE8FC}
\definecolor{dioBlueStroke}{HTML}{6C8EBF}

\definecolor{dioOrangeFill}{HTML}{FFE6CC}
\definecolor{dioOrangeStroke}{HTML}{D79B00}

\definecolor{dioRedFill}{HTML}{F8CECC}
\definecolor{dioRedStroke}{HTML}{B85450}

\definecolor{dioGreenFill}{HTML}{D5E8D4}
\definecolor{dioGreenStroke}{HTML}{82B366}

\definecolor{dioYellowFill}{HTML}{FFF2CC}
\definecolor{dioYellowStroke}{HTML}{D6B656}

\DeclareRobustCommand{\diobox}[3]{%
  \begingroup
  \setlength{\fboxsep}{1pt}
  \setlength{\fboxrule}{0.4pt}
  \fcolorbox{#1}{#2}{{#3}}%
  \endgroup
}

\DeclareRobustCommand{\bluebox}[1]{\diobox{dioBlueStroke}{dioBlueFill}{#1}}
\DeclareRobustCommand{\orangebox}[1]{\diobox{dioRedStroke}{dioOrangeFill}{#1}}
\DeclareRobustCommand{\redbox}[1]{\diobox{dioRedStroke}{dioRedFill}{#1}}
\DeclareRobustCommand{\greenbox}[1]{\diobox{dioGreenStroke}{dioGreenFill}{#1}}
\DeclareRobustCommand{\yellowbox}[1]{\diobox{dioYellowStroke}{dioYellowFill}{#1}}

\usepackage{listings}
\usepackage{xcolor}

\definecolor{codebg}{RGB}{248,248,248}      
\definecolor{codeborder}{RGB}{210,210,210}  
\definecolor{codekw}{RGB}{0,92,197}         
\definecolor{codestring}{RGB}{163,21,21}    
\definecolor{codecomment}{RGB}{120,120,120} 
\definecolor{codeid}{RGB}{45,45,45}         

\lstdefinestyle{medcbr}{
    language=Python,
    basicstyle=\ttfamily\footnotesize\color{codeid},
    keywordstyle=\color{codekw}\bfseries,
    stringstyle=\color{codestring},
    commentstyle=\color{codecomment}\itshape,
    backgroundcolor=\color{codebg},
    frame=single,
    rulecolor=\color{codeborder},
    framerule=0.4pt,
    xleftmargin=0pt,
    xrightmargin=0pt,
    framexleftmargin=4pt,
    framexrightmargin=4pt,
    framextopmargin=4pt,
    framexbottommargin=4pt,
    showstringspaces=false,
    tabsize=4,
    breaklines=true,
    breakatwhitespace=true,
    columns=fullflexible,
    captionpos=b
}

\definecolor{cvprblue}{rgb}{0.21,0.49,0.74}
\usepackage[pagebackref,breaklinks,colorlinks,allcolors=cvprblue]{hyperref}

\usepackage[capitalize]{cleveref}
\crefname{section}{Sec.}{Secs.}
\Crefname{section}{Section}{Sections}
\Crefname{table}{Table}{Tables}
\crefname{table}{Tab.}{Tabs.}

\def\paperID{44403}
\def\confName{CVPR}
\def\confYear{2026}

\begin{document}

\title{Vision-Language Models Encode Clinical Guidelines for \\Concept-Based Medical Reasoning}

\author{
Mohamed Harmanani$^{1,4}$, Bining Long$^{1*}$, Zhuoxin Guo$^{1,4*}$,  Paul F.R. Wilson$^{1,4}$, \\Amirhossein Sabour$^3$, Minh Nguyen Nhat To$^2$, Gabor Fichtinger$^1$, \\Purang Abolmaesumi$^{2\dagger}$, Parvin Mousavi$^{1,4\dagger}$ \\
$^1$Queen's University 
\quad$^2$University of British Columbia \quad$^3$McMaster University \quad$^4$Vector Institute \\
\small \texttt{mohamed.harmanani@queensu.ca}
}

\maketitle

\renewcommand{\thefootnote}{\fnsymbol{footnote}}
\footnotetext[1]{denotes equal contribution (with interchangeable order)}
\footnotetext[2]{denotes co-senior authorship}
\renewcommand{\thefootnote}{\arabic{footnote}}

\begin{abstract}
Concept Bottleneck Models (CBMs) are a prominent framework for interpretable AI that map learned visual features onto a set of meaningful concepts, to be used for task-specific downstream predictions. Their sequential structure enhances transparency by connecting model predictions to the underlying concepts that support them. In medical imaging, where transparency is essential, CBMs offer an appealing foundation for explainable model design. However, their discrete concept representations overlook broader clinical context such as diagnostic guidelines and expert heuristics, reducing reliability in complex cases. We propose MedCBR, a concept-based reasoning framework that integrates clinical guidelines with vision–language and reasoning models. Labeled clinical descriptors are transformed into guideline-conformant text, and a concept-based model is trained with a multi-task objective combining multi-modal contrastive alignment, concept supervision, and diagnostic classification to jointly ground image features, concepts, and pathology. A reasoning model then converts these predictions into structured clinical narratives that explain the diagnosis, emulating expert reasoning based on established guidelines. MedCBR achieves superior diagnostic and concept-level performance, with AUROCs of 94.2\% on ultrasound and 84.0\% on mammography. Further experiments were also performed on non-medical datasets, with 86.1\% accuracy. Our framework enhances interpretability and forms an end-to-end bridge from medical image analysis to decision-making.

\end{abstract}

\section{Introduction}
As the demand for transparent and explainable AI grows, Concept Bottleneck Models (CBMs) have become a widely studied approach~\cite{koh2020concept, yuksekgonul2022post, oikarinen2023label} for connecting model predictions to human-interpretable concepts. These models achieve interpretability by learning an intermediate layer of meaningful concepts, which are then mapped to task-specific outcomes through a separate predictive head. CBMs are widely used in computer vision tasks that benefit from concept-level prediction such as fine-grained recognition~\cite{xie2025discovering}, visual question answering~\cite{fu2023dynamic}, and scene understanding~\cite{barbiero2024relational} , but their appeal is particularly strong in medical imaging, where interpretability and trust are essential for clinical adoption~\cite{chowdhury2024adacbm, bunnell2024learning, pang2024integrating}.\\
\begin{figure}[t]
    \centering
    \includegraphics[width=\columnwidth]{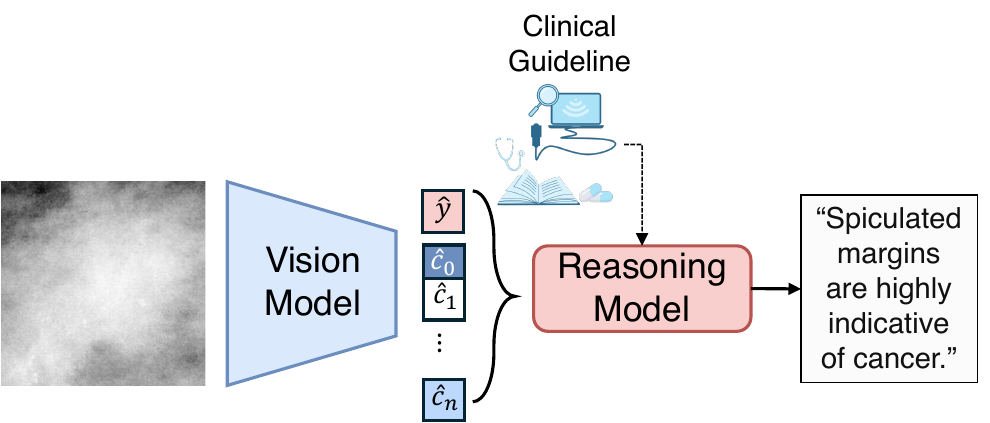}
    \caption{MedCBR: We frame interpretable medical image analysis as reasoning over diverse sources of evidence, including model predictions and clinical guidelines.}
    \label{fig:placeholder}
\end{figure}
\indent However, concept-based interpretability remains challenging in practice. In certain clinical tasks such as cancer detection, intermediate concepts often reflect risk assessments (e.g., BI-RADS categories for breast cancer) rather than definitive outcomes. These assessments may incorporate the clinician’s intuition or concern based on subtle cues in the image, even when individual features appear benign. Standard CBMs lack the capacity to capture this nuanced, experience-driven reasoning. Consequently, they may fail to identify cases that appear benign but raise concern when considered within the broader clinical context, thereby limiting performance in complex cases. Furthermore, concept annotations in medical datasets are often noisy or incomplete due to inter-observer variability and inherent ambiguity in imaging findings. This weakens the correspondence between visual evidence and concept labels, making it difficult for traditional CBMs to learn reliable medical representations. Bridging these gaps requires models that combine structured reasoning with domain knowledge by linking interpretable concepts and contextual guidelines within a unified framework. 

\indent To that end, we introduce MedCBR, a \textbf{med}ical \textbf{c}oncept-\textbf{b}ased \textbf{r}easoning framework for interpretable cancer detection. MedCBR explicitly models the structured clinical workflow used in diagnostic imaging, progressing from visual descriptors to high-level findings to diagnostic conclusions. By aligning model predictions with clinical reasoning, MedCBR enables both accurate and transparent decision-making, offering rich and structured text outputs that closely reflect how radiologists communicate uncertainty, risk, and next steps. In particular, we make the following contributions: 

1. We present a clinician-facing reasoning module that generates structured diagnostic narratives by integrating clinical guidelines with the predictions of a concept-based model to emulate the reasoning process used by clinicians, and produce transparent explanations of its predictions.

2. We develop a concept enrichment strategy that mitigates noise and incompleteness in human-annotated concepts by leveraging a large vision–language model (LVLM) to generate structured reports conditioned on the image, the concept ground truths, and the guideline. These enriched textual representations capture the contextual and relational meaning of visual findings, providing stronger and more consistent supervision.

3. We train a multi-task vision–language concept model that contrastively aligns images with the LVLM-enriched reports while jointly optimizing for concept and diagnosis prediction. This formulation encourages the vision encoder to learn clinically meaningful representations within a shared embedding space. Models trained under this strategy demonstrate improved generalization and higher diagnostic performance across multiple benchmarks.

\section{Related Work}

\textbf{Explainable AI.} A wide range of methods have been proposed to explain the decision-making process of neural networks. Classical approaches include parameter attribution methods~\cite{lundberg2017unified, ribeiro2016should}, gradient- and saliency-based visualizations~\cite{simonyan2013deep, selvaraju2017grad, smilkov2017smoothgrad, sundararajan2017axiomatic}, and attention-based models~\cite{abnar2020quantifying, darcet2023vision} that highlight influential image regions. Concept Bottleneck Models (CBMs)~\cite{koh2020concept} extend this idea by introducing structured, human-interpretable variables that mediate predictions~\cite{kim2018interpretability}, and have since become a central paradigm for interpretable AI across domains, including medical imaging~\cite{yan2023robust, wang2024concept, pang2024integrating, kim2023concept, bunnell2024learning}. Several extensions to the CBM framework have been proposed to move beyond the need for labelled concepts. These \emph{label-free} variants typically transform pre-trained backbones into CBMs by relying on LLM-generated concept vocabularies and image-text alignment objectives~\cite{yan2023learning,yuksekgonul2022post,oikarinen2023label,yang2023language}. AdaCBM~\cite{chowdhury2024adacbm} adapts this framework for medical imaging by introducing a learnable adapter between CLIP and the bottleneck to refines visual representations and mitigate domain shift. A limitation of label-free methods, however, is their reliance on automatically generated textual concepts, which can omit clinically important features or introduce spurious correlations. A complementary line of work aligns predictions with intermediate evidence using auxiliary objectives, such as enforcing diagnosis-evidence consistency with logical constraints~\cite{wang2021image}, or supervising attribution behavior with foreground masks~\cite{ben2024localization}. Much of the literature on intermediate concepts or auxiliary evidence improves interpretability by surfacing relevant clinical cues. However, tightly coupling standardized concept predictions with domain guidelines, especially in applications where concepts must adhere to strict definitions, remains less explored. To address this, we predict fine-grained clinical descriptors and condition the reasoning stage on them, yielding auditable and clinically verifiable decisions.\\
\indent\textbf{Integrating Clinical Knowledge.} Another class of explainable methods have explored the integration of clinical knowledge into LLM-based systems to improve transparency in medical imaging. Agentic approaches orchestrate tools and guidelines for interpretable diagnostic assistance. For instance, MAGDA~\cite{bani2024magda} proposes a guideline-driven agentic system for screening diagnosis. MedRAX~\cite{fallahpour2025medrax} couples chest X-ray foundation models with multi-modal LLMs to answer complex clinical queries using retrieval augmentation. Other works have sought to improve medical image analysis using reasoning~\cite{pan2025medvlm, huang2025medvlthinker} whilst others use clinical concepts to improve report generation~\cite{gu2025radalign, li2025reevalmed}. Besides structured concepts and clinical guidelines, another related direction incorporates knowledge graphs~\cite{abdullah2025vlm} to enhance radiological explanations~\cite{hamza2025llava}. However, prior work has focused on the use of concepts or guidelines as added context rather than explicit mechanisms to constrain concept-to-decision reasoning. We fill this gap by conditioning the reasoning process on fine-grained concept predictions, yielding verifiable and auditable reasoning from concepts and clinical guidelines.

\begin{figure*}[ht!]
    \centering    \includegraphics[width=\linewidth]{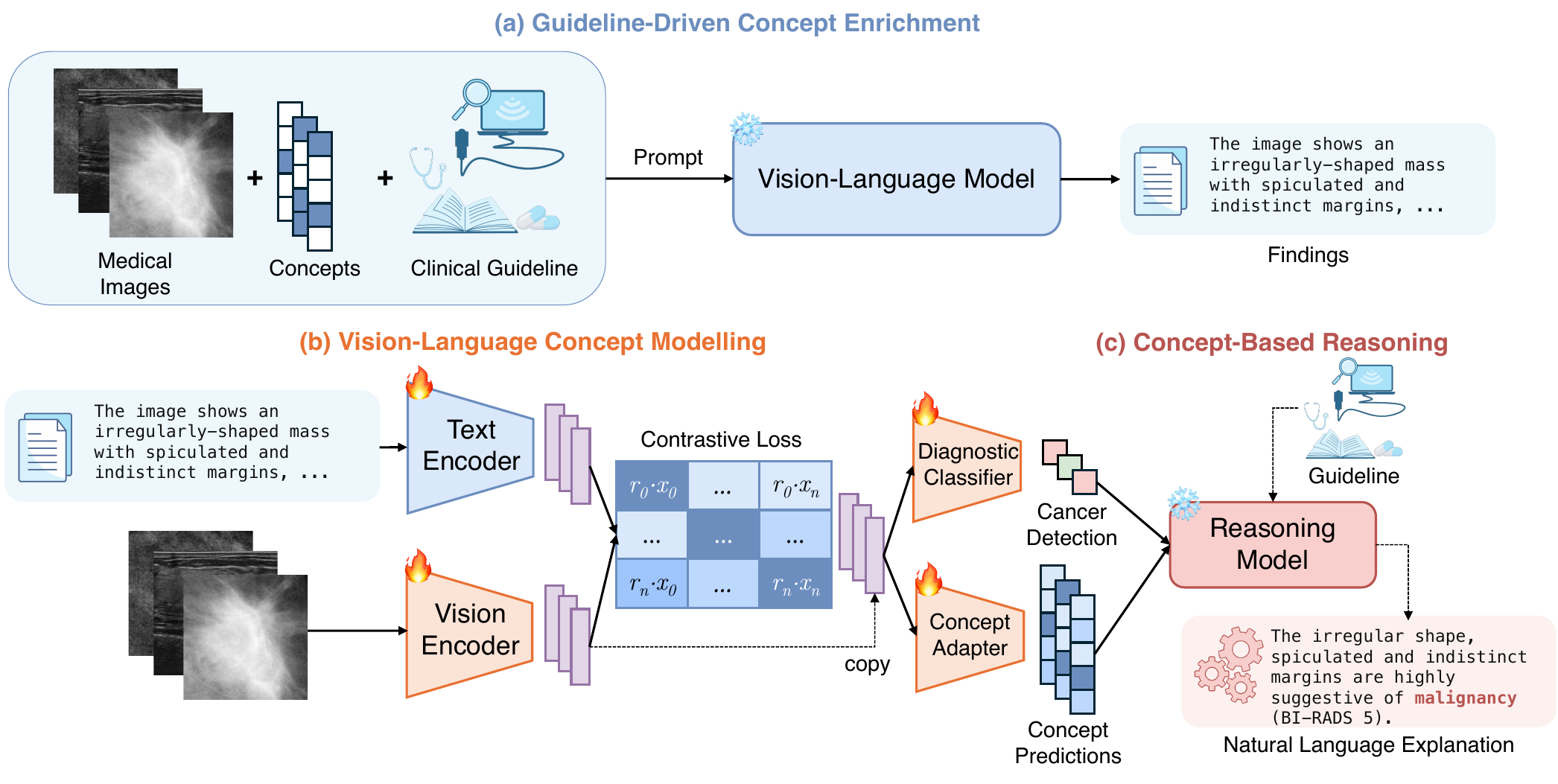}
    \caption{\textbf{Overview of the MedCBR framework.}
    MedCBR integrates clinical knowledge, vision--language alignment, and reasoning for interpretable cancer detection. 
    (a)~A large vision--language model (LVLM) generates guideline-conformant reports from input images and coarse concept annotations. 
    (b)~A concept-based CLIP model aligns visual and textual embeddings while jointly predicting concepts and diagnostic labels. 
    (c)~A frozen large reasoning model (LRM) synthesizes guideline information, predicted concepts, and model outputs to produce structured clinical explanations.}
    \label{fig:mainfig}
\end{figure*}

\section{Methods}
\subsection{Problem Formulation}

Let $\mathcal{D} = \{(x, c, y)\}$ denote a dataset of medical images $x \in \mathcal{X}$, concept annotations $c \in \{0,1\}^{N_c}$, and diagnostic labels $y \in \mathcal{Y} = \{\text{benign}, \text{malignant}\}$. 
Classical CBMs decompose the prediction $\hat{y}$ into a feature extractor $f : \mathcal{X} \rightarrow \mathbb{R}^d$, a concept predictor $g : \mathbb{R}^d \rightarrow [0,1]^{N_c}$, and a downstream classifier 
$q : [0,1]^{N_c} \rightarrow [0,1]$, yielding $\hat{y} = q(g(f(x)))$.
While this structure provides interpretability, it assumes that (i) the concept set $c$ is complete and noise-free, and (ii) diagnostic reasoning is a deterministic function of concept presence. 
In medical imaging, these assumptions are restrictive: concept annotations are often uncertain or incomplete, and real-world reasoning depends on contextual information encoded in clinical guidelines.\\
\indent MedCBR (Fig.~\ref{fig:mainfig}) addresses these limitations by modeling diagnosis as reasoning over multiple sources of evidence rather than a direct function of predicted concepts. 
Our framework jointly considers visual information from the image, interpretable concept predictions, and contextual knowledge from clinical guidelines to arrive at an explainable diagnostic conclusion. 
Formally, given a visual representation $h = f(x)$, the model predicts both a diagnostic label and an accompanying explanation:
\begin{align*}
\hat{y} = q(h), \quad \hat{c} = g(h), \quad 
\mathcal{E} = \mathcal{R}(\hat{y}, \hat{c}, \mathcal{G}),
\end{align*}
where $q : \mathbb{R}^d \rightarrow \{0,1\}$ maps image features to cancer predictions, 
$g : \mathbb{R}^d \rightarrow [0,1]^{N_c}$ infers interpretable clinical concepts, 
and $\mathcal{R}$ integrates these with the guideline $\mathcal{G}$ to produce the corresponding explanation.

\subsection{Guideline-Driven Concept Enrichment}

The discrete concept vector $c$ provides a structured yet limited description of image content: it indicates which findings are present but not how they relate or what they imply diagnostically. 
However, clinical interpretation, depends on relational and contextual meaning, specifically how combinations of findings align with reasoning patterns prescribed by diagnostic guidelines. 
To address this, we enrich $c$ into a continuous, guideline-conditioned representation that captures both the semantics and diagnostic context of the observed findings. 
Formally, we define an enrichment function $r = \mathcal{T}(x, c, \mathcal{G})$,
which fuses the visual information in $x$, the concept structure $c$, and the contextual knowledge encoded in $\mathcal{G}$ into a coherent textual representation $r$. 
We instantiate $\mathcal{T}$ using a pre-trained large vision–language model (LVLM) that generates structured reports from clinically defined concepts. 

For each image $x$, our goal is to obtain an enriched textual report that situates the set of exhibited concepts $c^{+} = \{c_j \mid c_j = 1,\forall c_j \in c \}$ within a clinically meaningful narrative. 
To achieve this, we prompt the LVLM with the image $x$, the positive concept set $c^{+}$, the label $y$, and the clinical guideline $\mathcal{G}$. 
The model is instructed to generate a report that (i) describes the visual findings corresponding to the provided concepts and (ii) summarizes their diagnostic implications according to $\mathcal{G}$. 
This process transforms the discrete concept vector $c$ into a clinically grounded textual representation $r$ that links observed evidence to diagnostic reasoning in a guideline-consistent manner.

\subsection{Vision--Language Concept Modelling}
Given the enriched reports generated in the previous stage, we train a vision–language model to align images and their textual explanations while preserving concept-level interpretability. 
We adopt CLIP~\cite{radford2021learning} as the backbone, consisting of a vision encoder $f^v_\theta(\cdot)$ and a text encoder $f^t_\phi(\cdot)$. 
CLIP’s contrastive formulation provides a natural inductive bias for aligning heterogeneous modalities: in the clinical setting, it enables the model to associate visual attributes (e.g., shape, margin) with their textual counterparts, grounding the visual embedding in interpretable semantics. 
For each image–report pair $(x, r)$, we compute embeddings $h_v = f^v_\theta(x)$ and $h_t = f^t_\phi(r)$ and apply a symmetric InfoNCE contrastive loss:
\begin{align*}
\mathcal{L}_{\text{CLIP}} = -\log \bigg(
\frac{\exp(\mathrm{sim}(h_v, h_t)/\tau)}
{\sum_{r'} \exp(\mathrm{sim}(h_v, h_t')/\tau)}\bigg),
\end{align*}
where $\mathrm{sim}(\cdot,\cdot)$ denotes cosine similarity and $\tau$ is a temperature parameter. 
Image–text alignment provides richer and more grounded supervision, reinforcing cross-modal consistency between visual and textual semantics. 

To incorporate clinical supervision, two auxiliary heads operate on the visual embedding $h_v$: 
a diagnostic classifier $W_y$ predicting the disease label $y$ via a linear projection, supervised with $
\mathcal{L}_{y} = \mathcal{L}_{\text{CE}}(W_y h_v, y)$,
and a concept prediction module $g = \{g_{\psi_i}\}_{i=1}^{N_c}$ that captures fine-grained attributes through $N_c$ specialized adapters, optimized by
\begin{align*}
\mathcal{L}_{c} = \frac{1}{N_c}\sum_{i=1}^{N_c} 
\mathcal{L}_{\text{CE}}(g_{\psi_i}(h_v), c_i).
\end{align*}
The diagnostic head promotes task-level separability, while the concept head encourages disentanglement along clinically meaningful axes of variation.
The final objective combines all terms:
\begin{align*}
\mathcal{L}_{\text{MedCBR}} =
\lambda \mathcal{L}_{\text{CLIP}} +
\mu \mathcal{L}_{y} +
\nu \mathcal{L}_{c},
\end{align*}
where $\lambda$, $\mu$, and $\nu$ control task weighting. 
Both encoders are initialized from pretrained CLIP checkpoints and fine-tuned jointly on $\mathcal{D}$. 
Concept adapters $g_{\psi_i}$ are implemented as lightweight two-layer MLPs.
This multi-task formulation jointly enforces (i) cross-modal alignment between image and report, (ii) concept-level interpretability, and (iii) discriminative power for diagnostic classification, yielding representations that are both semantically rich and clinically grounded.

\begin{table*}[ht!]
\caption{Quantitative classification results on Ultrasound and Mammography (5-fold cross-validation), with external validation on CUB-200 (holdout set). All results are averaged across multiple runs, with standard deviation shown. $^*$indicates reported accuracy.}

\centering
\setlength{\tabcolsep}{7pt}
\begin{tabular}{lcc|cc|c}
\toprule
\multirow{2}{*}{\bf Method} 
& \multicolumn{2}{c|}{BUS-BRA} 
& \multicolumn{2}{c|}{CBIS-DDSM} 
& \multicolumn{1}{c}{CUB-200-2011} \\
\cmidrule(lr){2-6}
& \bf AUROC  & \bf Bal. Accuracy  & \bf AUROC  & \bf Bal. Accuracy  & \bf Accuracy\\
\midrule
CLIP RN50 & 87.4 $\pm$1.4 & 80.9 $\pm$1.7 & 73.3 $\pm$1.7 & 67.7 $\pm$1.6 & 60.1 $\pm$1.1 \\ 
CLIP ViT-B/32 & 90.1 $\pm$1.5 & 83.1 $\pm$1.9 & 79.8 $\pm$0.9 & 72.3 $\pm$0.9 & 69.0 $\pm$0.4 \\ 
CLIP ViT-L/14 & \underline{93.5 $\pm$1.2} & \underline{88.1 $\pm$1.7} & \underline{82.4 $\pm$2.5} & \underline{75.5 $\pm$2.4} & \underline{85.7 $\pm$0.2} \\
SigLIP & 90.8 $\pm$0.8 & 85.0 $\pm$1.5 & 82.3 $\pm$1.4 & 74.6 $\pm$1.3 & 78.6 $\pm$0.6  \\
BiomedCLIP & 89.0 $\pm$0.5 & 82.1 $\pm$0.9 & 77.9 $\pm$0.5 & 71.1 $\pm$0.7 & --  \\
\midrule 
CBM~\cite{koh2020concept} & 84.8 $\pm$2.3 & 79.1 $\pm$2.0 & 79.6 $\pm$1.3 & 73.3 $\pm$1.3 &   62.9$^*$ \\
CLIP CBM~\cite{koh2020concept} & 91.8 $\pm$0.9 & 86.4 $\pm$1.7 & 81.8 $\pm$1.0 & 75.8 $\pm$0.9 &  67.0 $\pm$0.4  \\
P-CBM~\cite{yuksekgonul2022post} & 80.1 $\pm$0.5 & 73.9 $\pm$0.3 & 72.7 $\pm$0.0 & 67.0 $\pm$0.5 & 59.6$^*$\\
P-CBMh~\cite{yuksekgonul2022post} & 87.0 $\pm$1.7 & 75.0 $\pm$3.0 & 77.2 $\pm$1.8 & 69.3 $\pm$1.0 & 61.0$^*$\\
Label-free CBM~\cite{oikarinen2023label} & 60.0 $\pm$2.8 & 59.1 $\pm$2.1 & 70.0 $\pm$0.4 & 65.3 $\pm$0.5  & 74.3 $\pm$0.3$^*$ \\
AdaCBM~\cite{chowdhury2024adacbm} & 87.9 $\pm$0.5 & 80.5 $\pm$0.8 & 75.6 $\pm$3.0 & 68.9 $\pm$2.4 &  69.8 $\pm$0.2 \\
\midrule
\bf MedCBR & \bf 94.2 $\pm$0.4 & \bf 89.0 $\pm$0.9 & \bf{84.0 $\pm$0.7} & \bf{76.4 $\pm$0.6} &  \textbf{86.1 $\pm$0.2} \\
\bottomrule
\end{tabular}
\label{tab:perf_metrics}
\end{table*}

\subsection{Concept-Based Clinical Reasoning}
The final stage of our framework integrates clinical knowledge, visual evidence, and concept predictions into a coherent diagnostic rationale. 
While previous stages produce aligned vision–language representations, they do not explicitly model the reasoning process that links these observations to a diagnostic conclusion. 
To that end, we prompt a large reasoning model (LRM), conditioned on the outputs of the concept-based CLIP module, to explain its predictions. 
Let $\hat{y}=\sigma\big(f^y_\theta(h_v)\big)\in[0,1]$ denote the CLIP model’s predicted probability of malignancy and 
$\hat{c}=(\hat{c}_1,\dots,\hat{c}_{N_c})$ with $\hat{c}_i=\sigma\big(f^{c_i}_\theta(h_v)\big)$ the predicted concept probabilities. 
Given the relevant clinical guideline text $\mathcal{G}$, the LRM $\mathcal{R}$ receives a structured prompt $\pi$ comprising $(\hat{y},\hat{c},\mathcal{G})$ and produces a guideline-referenced explanation: $\mathcal{E} = \mathcal{R}(\pi)$.

\textbf{Prompt Structure.} Each prompt $\pi$ follows a fixed template consisting of four components:
(i) a brief task instruction,
(ii) the model’s cancer prediction $\hat{y}$,
(iii) the list of predicted concepts and their associated confidences, and
(iv) the relevant section of the BI-RADS guideline $\mathcal{G}$ defining relationships between clinical descriptors and malignancy risk.
The LRM is instructed to first interpret how each concept contributes to the final decision, then cross-check the reasoning against $\mathcal{G}$, and finally provide a concise, step-by-step justification of the outcome.
Formally, $\pi$ can be represented as a concatenation $\pi = \big( \mathcal{Q},\, \hat{y},\, \hat{c},\, \mathcal{G} \big)$, 
where $\mathcal{Q}$ specifies the reasoning task, $\hat{y}$ and $\hat{c}$ denote the model’s diagnostic and concept-level outputs, and $\mathcal{G}$ provides the relevant clinical guideline for context.

\textbf{Grounded and Reliable Reasoning.}
Because the LRM operates over a structured prompt $\pi$ and is explicitly conditioned on $\mathcal{G}$, its reasoning remains anchored to verifiable clinical knowledge rather than unconstrained text generation.
The inclusion of model-derived predictions $(\hat{y}, \hat{c})$ and standardized guideline excerpts constrains the search space of possible outputs, reducing the likelihood of unsupported or speculative statements, minimizing hallucination risk.
\begin{table}[t]
\centering
\footnotesize
\caption{Ablation on model components. Performance improves as concept and text supervision are successively introduced.}
\label{tab:ablation1}
\begin{tabular}{lccc}
\toprule
\textbf{Variant} & \textbf{BUS-BRA} & \textbf{CBIS-DDSM} & \textbf{CUB-200} \\
\midrule
CLIP ViT & 93.5 & 82.4 & \underline{85.7} \\
\midrule
CLIP+CBL & 91.8 & 81.8 & 67.0 \\
\quad +Guideline & 92.0$^{\up{0.2}}$ & 83.1$^{\up{1.3}}$ & {72.9}$^{\up{5.0}}$ \\
\midrule
CLIP+MTL & \underline{93.6} & \underline{83.2} & 82.3 \\
\quad +Guideline & \textbf{94.2}$^{\up{0.8}}$ & \textbf{84.0}$^{\up{0.8}}$ & \textbf{86.1}$^{\up{3.8}}$ \\
\bottomrule
\end{tabular}
\end{table}
\section{Experiments}
\subsection{Setup}

\noindent\textbf{Datasets.} We train and evaluate our models on three primary datasets spanning medical and natural images. 
For breast ultrasound, we use BrEaST~\cite{pawlowska2024curated} and BUS-BRA~\cite{gomez2024bus}. BUS-BRA contains 1,875 images from 1,064 patients with biopsy-proven pathology and BI-RADS 2–5 assessments, and is used for cross-validation-based training, validation, and testing. BrEaST contains a total of 256 expert-annotated images, and is used to augment the training data. For mammography, we use CBIS-DDSM~\cite{sawyer2016curated}, a widely used benchmark for breast cancer detection containing 10,239 mammography images curated from the Digital Database for Screening Mammography with lesion annotations and verified pathology. For external validation beyond the medical domain, we use CUB-200-2011~\cite{WahCUB_200_2011}, which contains 11,788 images across 200 bird species with part locations and 312 concept annotations. Following prior work, we used the 112 most prominent concepts~\cite{koh2020concept,yuksekgonul2022post}. To construct the guideline for each dataset, we used ChatGPT~\cite{openai2023gpt4} to condense the \textit{Sibley Field Guide to Birds}~\cite{sibley2022sibley} for CUB-200 and paraphrase the relevant BI-RADS Atlas sections~\cite{magny2023breast} for BUS-BRA and CBIS-DDSM.\\
\begin{figure*}[ht!]
    \centering
    \includegraphics[width=0.95\textwidth]{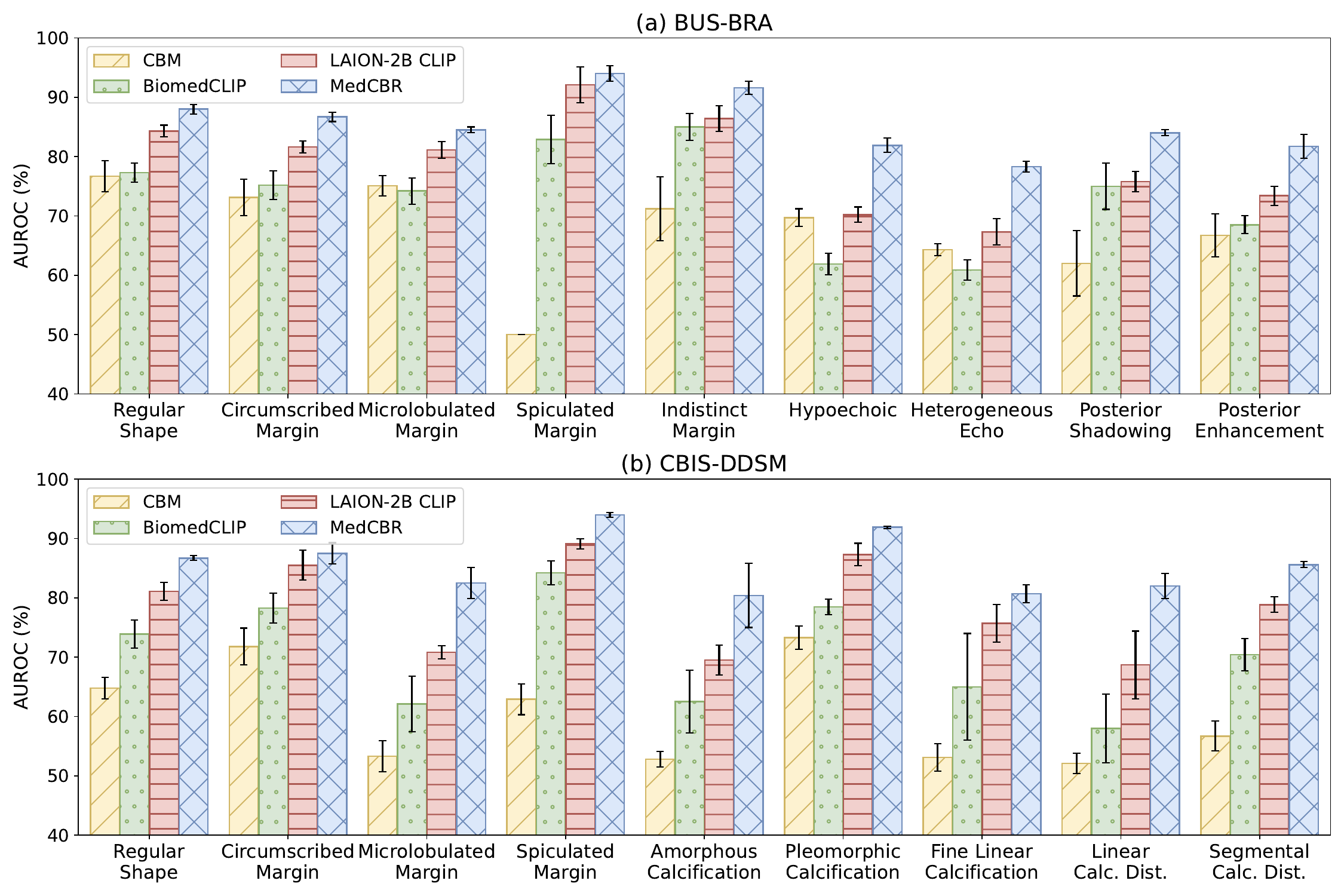}
    \caption{Quantitative results on concept detection for ultrasound and mammography. For each dataset, we focus on a subset of concepts most relevant to cancer detection.}
    \label{fig:concept_analysis}
\end{figure*}
\indent\textbf{Baselines.} To evaluate our method, we benchmark it against several prominent baselines in the literature: \\
\textbullet~\textbf{CBM}~\cite{koh2020concept}: the original concept bottleneck model.\\ 
\textbullet~\textbf{PCBM}~\cite{yuksekgonul2022post}: post-hoc CBM that learns a linear classifier over concept activations extracted from a frozen backbone.\\
\textbullet~\textbf{Label-free CBM}~\cite{oikarinen2023label}: discovers latent concept representations on top of CLIP without explicit concept supervision.
\textbullet~\textbf{AdaCBM}~\cite{chowdhury2024adacbm}: an adaptive label-free CBM that leverages a learnable adapter to improve medical classification. \\
\textbullet~\textbf{CLIP}~\cite{radford2021learning}: strong vision encoders pre-trained on language-image alignment. We evaluate OpenAI CLIP RN50 and ViT-B variants~\cite{radford2021learning}, a CLIP ViT-L/14 pretrained on LAION-2B~\cite{schuhmann2022laion}, a SigLIP ViT-B/16 pretrained on WebLI~\cite{tschannen2025siglip}, and BiomedCLIP~\cite{zhang2023biomedclip}.\\
\textbullet~\textbf{Off-the-shelf VLMs}: zero-shot prompt-based VLMs such as Qwen2.5VL-7B~\cite{bai2025qwen2} and MedGemma-4B~\cite{sellergren2025medgemma}, a reference baseline for medical image--text understanding.\\
\indent\textbf{Evaluation. } 
We evaluate MedCBR on three aspects: classification performance, concept-level performance, and reasoning quality. 
For BUS-BRA and CBIS-DDSM, we report the AUROC and Balanced Accuracy due to their robustness to class imbalance. For CUB-200, we follow the literature and use accuracy. For concept evaluation, we compute per-concept AUROC. Finally, to assess reasoning quality, we use a combination of qualitative case studies as well as a structured evaluation rubric\footnote{The full rubric can be found in the Appendix.} developed by a trained radiologist to measure the coherence and clinical validity of the model’s explanations (Table~\ref{tab:clinical_utils}). 
The rubric includes three metrics: (i) the \textit{Concept Interpretation Score (CIntS)}, which measures whether each concept is interpreted correctly; (ii) the \textit{Concept Integration Score (CIgS)}, which assesses whether multiple concepts are combined coherently and conflicts between benign and malignant features are resolved; and (iii) the \textit{BI-RADS Assignment Score (BAS)}, which evaluates whether the final diagnostic category is assigned according to the correct guideline-based criteria given the observed features. A radiologist then evaluated 20 randomly selected cases while being blinded to the ground truth pathology and the composition of the evaluation cohort. The test samples were all drawn by a separate investigator to ensure independence between case selection and scoring. We further calculated the model's clinical utility using sensitivity, specificity, and F1 score.\\
\indent\textbf{Implementation details.} All experiments were run with PyTorch 2.1 on a single NVIDIA L40S GPU (24 GB). We used AdamW with an initial learning rate of $1\times10^{-5}$, cosine annealing over 150 epochs, and a 10-epoch warmup. Images were cropped to lesion ROIs, resized to $224\times 224$, and augmented with random translation, rotation, and flip. 

\subsection{Results}
\textbf{Diagnostic Performance.}
Table~\ref{tab:perf_metrics} summarizes the quantitative results for cancer detection and external validation. Across all datasets, MedCBR achieves strong performance, matching the strongest vision encoders while substantially outperforming concept-based models. On BUS-BRA, MedCBR attains the highest AUROC (94.2\%) and balanced accuracy (89.0\%), surpassing both CLIP ViT-L/14 (+0.7\%) and AdaCBM (+6.3\%). Similarly, on CBIS-DDSM, it outperforms concept-based baselines like CBM (+4.4\%) and PCBM-h (+6.7\%) and remains competitive with large-scale CLIP models. Across both datasets, MedCBR consistently outperforms both interpretable and black-box vision models, achieving the strongest performance without sacrificing transparency. The improvement over CBM variants highlights the complementary strengths of its design. The gain over classical CBMs and PCBM-h indicates that replacing the bottleneck layer with multi-task supervision grounded in guidelines enhances contextual understanding.
Furthermore, the improvement over AdaCBM, which adapts CLIP to medical data, demonstrates that domain grounding provides a stronger inductive bias over adapter fine-tuning alone.
\begin{figure*}[ht!]
    \centering
    \includegraphics[width=0.9\textwidth]{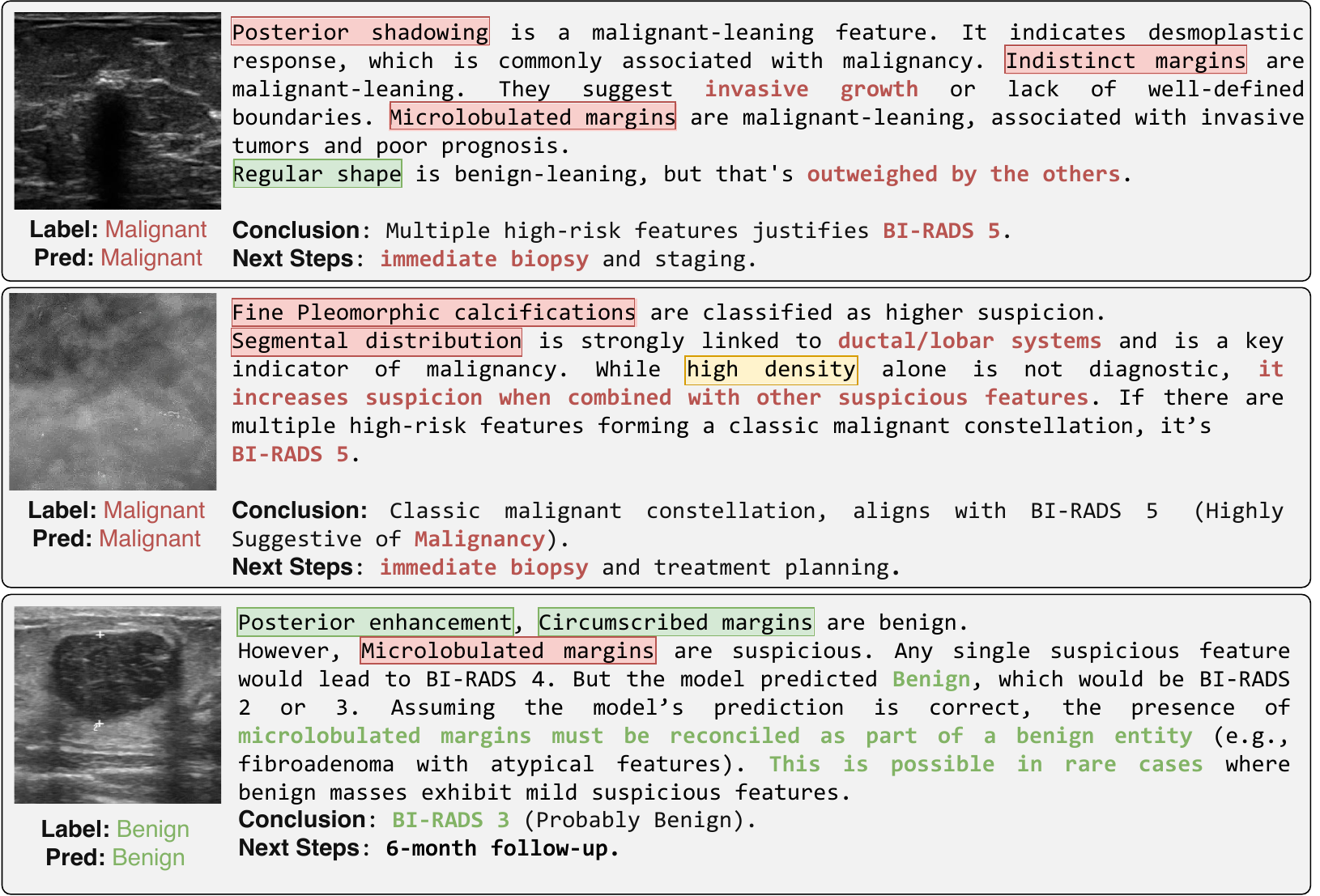}
    \caption{Case study of MedCBR's reasoning on medical images. Clinical concepts highlighted in \greenbox{green} support a benign conclusion, those in \redbox{red} support cancer, and those in \yellowbox{yellow} are neutral. Key observations are emphasized in \textbf{\textcolor{dioGreenStroke}{green}} (benign) and \textbf{\textcolor{dioRedStroke}{red}} (cancer).}
    \label{fig:reasoning_analysis}
\end{figure*}
Finally, on CUB-200-2011, MedCBR achieves 86.1\% accuracy, outperforming the label-free CBM by over 10\%. This underscores the limitations of unguided concept discovery and highlights the benefit of coupling concepts with domain context. Overall, our framework for grounding representations in domain guidelines has shown strong generalization capabilities in ultrasound, mammography, and natural images.\\
\indent\textbf{Ablation Study.}
Table~\ref{tab:ablation1} summarizes how each architectural component contributes to performance. We begin with removing the text branch entirely and introduce a Concept Bottleneck Layer (CBL), enforcing a deterministic mapping $y = W_y \cdot g(h_v)$ from predicted concepts to diagnosis, and leading to reduced performance. When CBL is trained with guideline-based reports (CBL+Guideline), performance improves, suggesting that multi-modal supervision helps structure the concept space in semantically meaningful ways. Finally, the multi-task formulation (MTL+Guideline, i.e., MedCBR) jointly optimizes diagnostic and concept objectives while maintaining contrastive alignment, achieving the highest performance across all datasets. These results show that integrating contrastive, concept-level, and diagnostic supervision yields representations that are both robust and semantically expressive.\\
\indent\textbf{Concept-Level Performance.} Figure~\ref{fig:concept_analysis} compares MedCBR’s concept detection performance against CBM, BiomedCLIP, and LAION-2B CLIP. Across both datasets, MedCBR achieves the highest concept AUROC, due to its multi-modal supervision, which allows it to capture discriminative and clinically meaningful feature representations. In contrast, CBMs show lower performance across most concept groups in our setting, likely due to their sole reliance on concept-label supervision and lack of pre-training. BiomedCLIP improves over CBM due to its domain-specific pretraining, but is outperformed by LAION-2B CLIP which benefits from broader web-scale vision–language pretraining. However, the latter's performance declines on modality-specific cues such as echogenicity and posterior features in ultrasound, and calcifications in mammography. We hypothesize that these phenomena depend on modality-specific imaging physics, and are thus under-represented in natural image corpora. Finally, MedCBR achieves consistently higher concept detection across categories and datasets. Its integration of multi-modal supervision allows it to capture both visually grounded and modality-specific features, indicating that clinically meaningful representations can emerge when visual, conceptual, and contextual cues are learned jointly.\\
\begin{table*}[t]
\caption{Evaluation of clinical utility against general-purpose and medical VLMs on BUS-BRA (left) / CBIS-DDSM (right). \newline($\mathcal{I}$: image, $\mathcal{G}$: guideline, $\hat{c}$: vision-predicted concepts.)}
\centering
\setlength{\tabcolsep}{5pt} 
\begin{tabular}{l|c|ccc|ccc}
\toprule
 \multirow{2}{*}{\bf Method} & \multirow{2}{*}{\bf $\mathcal{I,G},\hat{c}$} &
 {\bf Sens.} (\%) & {\bf Spec.} (\%) & {\bf F1} (\%) &
 {\bf CIntS}  (\%) & {\bf CIgS} (\%) & {\bf BAS.} (\%) \\
 & & \multicolumn{3}{c|}{$n = 379$ / $n=721$} & \multicolumn{3}{c}{$n = 20$} \\
\midrule
Radiologist & -- & 93.9 / 90.8 & 75.8 / 52.3 & 75.1 / 70.9 & -- & -- & -- \\
\midrule
Llama 3.2-11B & $\checkmark,\checkmark,\times$ & 33.0 / 0.00 & 70.0 / 58.0 & 32.6 / 0.00 & 17.5 / 0.00 & 15.0 / 0.00 & 0.00 / 0.00 \\
Qwen2.5VL-7B         & $\checkmark,\checkmark,\times$ & 30.0 / 42.0 & 0.00 / 53.0 & 16.6 / 40.7 & {\bf 100} / {\bf 100} & {\bf 100} / {\underline{95.0}} & {\underline {98.0}} / \underline{90.0} \\
MedGemma-4B         & $\checkmark,\checkmark,\times$ & 35.0 / 45.0 & 85.0 / {\bf 100} & 41.2 / 62.1 & {\bf 100} / {\bf 100} & {\bf 100} / {\bf 100} & {\bf 99.0} / {\bf 100} \\
CLIP+Qwen3-8B   & $\checkmark,\times,\checkmark$ & {\bf 82.5} / \underline{61.5} & \underline{92.5} / {\underline{87.5}} & {\underline{82.5}} / \underline{68.9} & \underline{98.4} / 93.9 & 86.3 / 80.0 & 73.0 / 65.0 \\
\midrule
{\bf MedCBR-8B} & $\checkmark,\checkmark,\checkmark$ &
{\bf 82.5} / {\bf 78.7} & {\bf 94.3} / {{76.0}} & {\bf 84.3} / {\bf 74.4} &
95.4 / \underline{95.2} & \underline{98.3} / 87.5 & {86.0} / 80.0 \\
\bottomrule
\end{tabular}
\label{tab:clinical_utils}
\end{table*}
\indent\textbf{Reasoning Quality.} Table~\ref{tab:clinical_utils} summarizes clinical utility and decision-making performance across VLM baselines. Most off-the-shelf VLMs attain near-ceiling guideline conformity scores, yet their sensitivity and F1 remain substantially lower. This suggests that while they followed guideline logic fluently, their rationales were often inconsistent with the actual visual evidence. We hypothesize that weaker grounding can coincide with higher coherence, especially when VLMs invoke the concepts that yield the most straightforward explanation, even when they are not well supported by the image, thereby circumventing the need to reconcile ambiguous or conflicting visual cues. Notably, MedGemma achieves a high specificity and the best F1 among the VLM baselines, although its sensitivity remains low. CLIP+Qwen3 provides a more grounded baseline by conditioning on vision-predicted concepts, but its reasoning competence scores are lower because it lacks explicit guideline structure and must operate on imperfect evidence.\\
\indent MedCBR achieves the highest overall clinical utility on both BUS-BRA and CBIS-DDSM, attaining the best F1 while maintaining high sensitivity and specificity. Compared to radiologists, MedCBR achieves markedly higher specificity on both datasets, indicating fewer false positives, and it also attains the highest sensitivity among the VLM baselines. These gains are illustrated by the model’s behavior in Fig.~\ref{fig:reasoning_analysis}: MedCBR produces coherent reasoning and decision patterns that mirror how radiologists weigh conflicting evidence, correctly interpreting malignant-leaning descriptors and linking them to plausible pathological correlates such as invasive growth (Fig.~\ref{fig:reasoning_analysis}a). The model also demonstrates the ability to integrate multiple features coherently. When several malignant findings are present, MedCBR correctly links them into a diagnostic constellation consistent with the appropriate BI-RADS risk score of 5 (Fig.~\ref{fig:reasoning_analysis}b). Remarkably, in this case, the radiologist’s risk assessment was BI-RADS~3 (i.e. probably benign), indicating that the model’s reasoning matched the underlying pathology more closely. When confronted with a mix of benign and suspicious cues, MedCBR appropriately weighs competing evidence and arrives at a clinically plausible and contextually justified conclusion. This is exemplified in atypical cases such as Fig.~\ref{fig:reasoning_analysis}c, where despite identifying microlobulated margins, the model recognizes that the co-occurrence of less suspicious features is more consistent with a benign fibroadenoma. This hypothesis was confirmed by pathology, indicating that the model captures how descriptor constellations relate to the final diagnosis.\\
\begin{figure}[t!]
    \centering
    \includegraphics[width=\columnwidth]{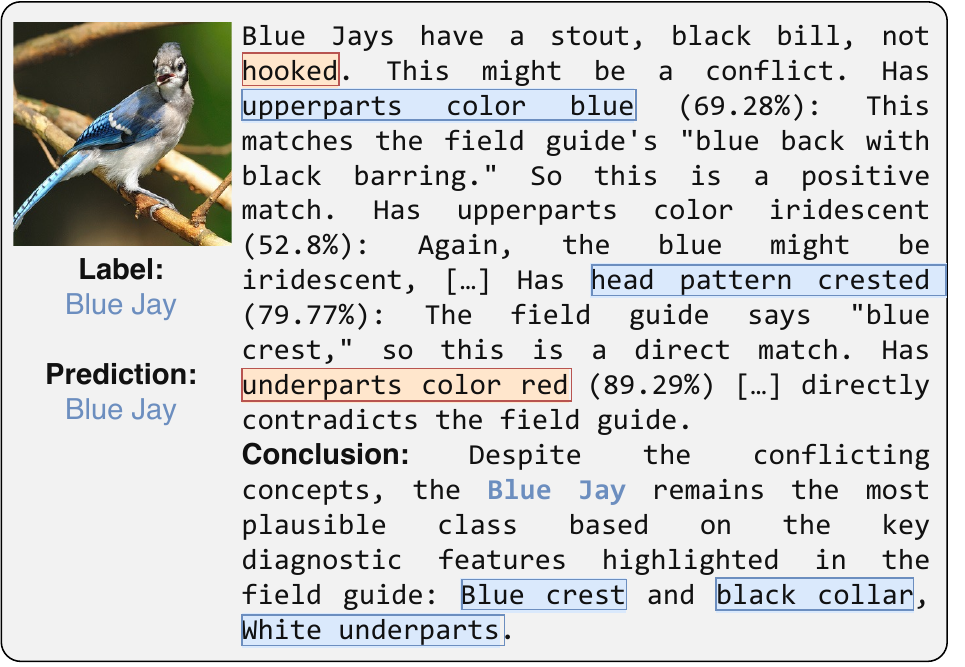}
    \caption{Validation of MedCBR's reasoning on CUB-200. Observations matching the guideline are highlighted in \bluebox{blue}, conflicting ones are highlighted in \orangebox{orange}.}
    \label{fig:birb_reason}
\end{figure}
\indent Interestingly, on CUB-200 (Fig.~\ref{fig:birb_reason}), the LRM also demonstrates an ability to challenge the concept model's predictions, explicitly noting inconsistencies between predicted attributes and the guideline. This suggests that the model is capable of reconciling conflicting inputs through rule-based reasoning grounded in the guideline, highlighting a degree of awareness absent from conventional CBMs.

\section{Conclusion}
\textbf{Reasoning Improves Transparency.}  
By modeling the reasoning process that connects clinical guidelines, concepts, and diagnosis, MedCBR provides radiologist-like image interpretation with coherent risk assessments and next-step recommendations. A guideline-conditioned reasoning module integrates model predictions with clinical rules, articulates how specific imaging concepts map to diagnostic implications and underlying mechanisms, and relates each presentation to canonical patterns described in guidelines and representative pathology cases. In short, MedCBR explains why \textit{specific} concepts matter and how \textit{combinations} of concepts alter risk, yielding reliable clinical reasoning.\\
\indent\textbf{Limitations and Future Work.} A limitation of our approach is the reliance on annotated concept labels, which are costly to obtain at scale. The development of reliable methods for label-free reasoning and concept modeling should thus be the subject of immediate future work. It is also worth investigating how this framework can be adapted for clinical deployment.





{\small
\bibliographystyle{ieee_fullname}

}


\newpage

\noindent{\Large \bf Appendix}

\section*{A. Additional Experimental Details}
\subsection*{A1. Code and configuration files}
Our code and configuration files are publicly available at: \href{https://www.github.com/mharmanani/medcbr}{github.com/mharmanani/medcbr}.

\subsection*{A2. Datasets and Concepts}

\textbf{BUS-BRA}~\cite{gomez2024bus} is a publicly available breast ultrasound dataset containing 1,875 images from 1,064 patients with biopsy-proven pathology labels and BI-RADS 2--5 assessments. Images were acquired retrospectively at the University of Brasília using a variety of scanners and linear-array transducers (5--14~MHz), resulting in heterogeneous imaging conditions characteristic of routine clinical practice. Lesions include masses abnormalities and are accompanied by radiologist-verified segmentation masks. All metadata were de-identified and curated to remove duplicates and low-quality scans. Dataset splits are constructed at the patient level to avoid data leakage. For this study, a trained radiologist annotated each lesion with 15 clinically relevant BI-RADS concepts to support concept-based modeling. The 15 concepts are shown in Table~\ref{tab:concepts_busbra}.

\begin{table}[h!]
    \centering
    \small
    \begin{tabular}{l|c}
    \toprule
        \bf Concept Category & \bf Concepts \\
    \midrule
        Posterior features & 
        (1) shadowing;\ (2) enhancement \\[0.25em]
        
        Other findings & 
        (3) halo;\ (4) skin thickening;\\
        & (5) calcifications \\[0.25em]
        
        Margins & 
        (6) circumscribed;\ (7) indistinct;\\
        & (8) angular;\ (9) microlobulated;\\
        & (10) spiculated \\[0.25em]
        
        Shape & 
        (11) regular \\[0.25em]
        
        Echogenicity & 
        (12) hypoechoic;\ (13) hyperechoic;\\
        & (14) heterogeneous;\ (15) cystic \\
    \bottomrule
    \end{tabular}
    \caption{Clinically annotated BI-RADS concepts used for concept-based modeling.}
    \label{tab:concepts_busbra}
\end{table}
\begin{table}[t!]
    \centering
    \small
    \begin{tabular}{l|c}
    \toprule
        \bf Concept Category & \bf Concepts \\
    \midrule
        Mass Shape &
        (1) regular (round/oval); 
        (2) irregular; \\
        & (3) lobulated \\[0.35em]

        Mass Margins &
        (4) circumscribed; 
        (5) ill-defined; \\
        &(6) spiculated;
        (7) obscured; \\
        &(8) microlobulated \\[0.35em]

        Calcification &
        (9) pleomorphic; 
        (10) amorphous; \\
        & (11) fine linear;
        (12) branching; \\
        & (13) vascular; 
        (14) coarse; \\
        & (15) punctate; 
        (16) lucent-centered; \\
        & (17) eggshell;
        (18) round; 
        (19) regular; \\
        & (20) dystrophic \\[0.35em]

        Calcif. Distrib. &
        (21) clustered; 
        (22) segmental; \\
        & (23) linear;
        (24) scattered; \\
        & (25) regional \\[0.35em]

        Breast Density &
        (26) low density \\
        & (27) moderate density\\
        & (28) high density  \\[0.35em]

        Other Findings &
        (29) architectural distortion; \\
        &(30) asymmetry;
        (31) lymph node \\
    \bottomrule
    \end{tabular}
    \caption{Ground-truth concept labels for CBIS-DDSM.}
    \label{tab:concepts_ddsm}
\end{table}
\begin{figure*}[t]
\centering
\small

\begin{lstlisting}[style=medcbr]
BIRADS_REPORTING_GUIDELINE_US = """
SUCCINCT DESCRIPTION OF THE OVERALL BREAST COMPOSITION (screening only)

Tissue composition patterns can be estimated more easily in the large FOVs of automated US scans but can also be discerned in the small FOV of a handheld US scan. The three US descriptors for tissue composition described earlier in the US lexicon, 'homogeneous
background echotexture-fat,' 'homogeneous background echotexture-fibroglandular,'
and 'heterogeneous background echotexture' (Table 3) (below) correspond loosely to
the four density descriptors of mammography and the four fibroglandular tissue descriptors of MRI. At US, breast tissue composition is determined by echogenicity. Subcutaneous fat, the tissue relative to which echogenicity is compared, is medium gray and darker
than fibroglandular tissue, which is light gray. Heterogeneous breasts show an admixture
of hypoechoic and more echogenic areas. Careful real-time scanning will help differentiate a small hypoechoic area of normal tissue from a mass.

Table 3. Breast Tissue
Tissue Composition
a. Homogeneous background echotexture-fat
b. Homogeneous background echotexture-fibroglandular
c. Heterogeneous background echotexture

CLEAR DESCRIPTION OF ANY IMPORTANT FINDINGS

The description of important findings should be made, in order of clinical relevance, using
lexicon terminology, and should include: [...]
"""
\end{lstlisting}

\begin{lstlisting}[style=medcbr]
BIRADS_DIAGNOSTIC_GUIDELINE_US = """
# BI-RADS Ultrasound Diagnostic & Stratification Guideline

## Scope & Purpose
Standardizes description and assessment of breast findings on ultrasound (US).  
Descriptors (shape, orientation, margins, echo pattern, posterior features, associated findings) are integrated to assign a BI-RADS category communicating malignancy likelihood.  
The most suspicious ('dominant') feature determines the category.

---

## 1. Mass: Shape
- **Oval / Round / Gently Lobulated** - favors benignity (e.g., fibroadenoma).
- **Irregular** - suspicious; upgrade to BI-RADS 4-5 depending on context.

## 2. Mass: Orientation
- **Parallel (Wider-than-tall)** - benign-leaning.
- **Not Parallel (Taller-than-wide)** - malignant-leaning; indicates tissue plane invasion.

## 3. Mass: Margins
- **Circumscribed** - benign (e.g., fibroadenoma, cyst).
- **Microlobulated / Indistinct / Angular / Spiculated** - suspicious for invasion. [...]
"""
\end{lstlisting}

\caption{Example guideline snippets used to condition the LVLM (reporting)
and the LRM (diagnostic).}
\label{fig:guideline_snippets}
\end{figure*}
\begin{figure*}[t]
\centering
\small

\begin{lstlisting}[style=medcbr]
{
    "Pomarine Jaeger": """
    SPECIES ID: Pomarine Jaeger (Stercorarius pomarinus)  
    OVERALL IMPRESSION: Largest jaeger; bulky with heavy chest and broad wings; breeding adults with twisted spoon-shaped tail projections.  
    KEY IDENTIFICATION FEATURES: Head: Dark cap; pale nape; heavy bill. Chest: Whitish to buffy chest with dark sides. Back/Wings: Dark back; broad wings with pale bases to primaries. Tail: Two thick twisted tail spoons in breeding adults. Legs/Feet: Dark.  
    BEHAVIOR AND POSTURE: Powerful, aggressive flier; pirates food from gulls/terns; hunts lemmings on tundra.  
    HABITAT CONTEXT: High Arctic tundra for breeding; winters offshore worldwide.  
    SIMILAR SPECIES: Parasitic Jaeger smaller; Long-tailed Jaeger slimmer with long streamers.  
    DIAGNOSTIC FIELD MARKS: Large bulky build, twisted tail spoons in breeding, heavy flight. """,      
    
    "Blue Jay": """SPECIES ID: Blue Jay (Cyanocitta cristata) 
    OVERALL IMPRESSION: Bold, noisy corvid; blue above, white below, black necklace; common in eastern North America.  
    KEY IDENTIFICATION FEATURES: Head: Blue crest; black collar; stout black bill. Chest: White chest and belly. Back/Wings: Blue back with black barring; white wing patches. Tail: Long blue tail with black bars and white tips. Legs/Feet: Dark.  
    BEHAVIOR AND POSTURE: Noisy calls; mimics hawks; caches acorns; social at feeders.  
    HABITAT CONTEXT: Woodlands, parks, suburbs in eastern North America.  
    SIMILAR SPECIES: Florida Jay, Western Scrub-Jay; Blue Jay larger with crest.  
    DIAGNOSTIC FIELD MARKS: Blue crest, black necklace, barred wings and tail, noisy calls. """, [...]
}
\end{lstlisting}
\caption{Example snippets of the field guide used to prompt the LVLM and LRM for CUB-200.}
\label{fig:birb_snippets}
\end{figure*}

\textbf{BrEaST}~\cite{pawlowska2024curated} is a high-quality breast ultrasound dataset containing 256 expertly annotated B-mode images with pixel-level lesion masks and BI-RADS descriptors. Images originate from multiple clinical settings and were curated to ensure consistent image quality, standardized acquisition views, and reliable expert annotations. In this work, BrEaST is used to augment BUS-BRA by providing additional training examples. Only images with complete annotations and verified pathology labels are included. Concept labels were provided by the authors of the dataset, and are the same as the ones highlighted in Table~\ref{tab:concepts_busbra}.

\indent\textbf{CBIS-DDSM}~\cite{sawyer2016curated} is a curated subset of the Digital Database for Screening Mammography and provides ROI-level patches for benign and malignant lesions, each paired with pathology-confirmed labels. The dataset includes calcification and mass subsets with standardized training and test partitions. For each ROI, structured clinical concepts such as margin, shape, breast density, calcification shape, and calcification distribution are available. Only cropped ROIs are used in this work, and the mass and calcification subsets were merged to create a single unified dataset of 3500 images. As done in BUS-BRA, dataset splits for each fold were constructed at the patient level. \\
\indent\textbf{CUB-200-2011}~\cite{WahCUB_200_2011}  is a fine-grained natural image dataset containing 11{,}788 bird images spanning 200 species. Each image is annotated with 312 binary part- and appearance-based attributes covering shape, color, and texture. We adopt the standard train/test split provided by the dataset's creators and use a reduced concept bank of 112 concepts, as is commonly done in the literature~\cite{koh2020concept, yuksekgonul2022post, oikarinen2023label}. This dataset serves as a natural image benchmark to evaluate the generality of MedCBR’s concept grounding and reasoning beyond medical imaging.

\section*{B. Integrating Domain Guidelines}
In this section, we provide further details on the construction of domain guidelines and their integration with large language and reasoning models. The full guideline text is available in our code. 

\subsection*{B1. BI-RADS Clinical Guideline}
For breast cancer detection, we constructed the following domain guidelines to support synthetic report generation and clinical reasoning:
\begin{enumerate}
    \item \texttt{BIRADS\_REPORTING\_GUIDELINE\_US}, 
    \item \texttt{BIRADS\_REPORTING\_GUIDELINE\_MG}, 
    \item \texttt{BIRADS\_DIAGNOSTIC\_GUIDELINE\_US},  
    \item \texttt{BIRADS\_DIAGNOSTIC\_GUIDELINE\_MG},  
\end{enumerate}

The two \emph{reporting} guidelines describe how to structure a BI-RADS-compliant clinical report from ultrasound and mammography images respectively, including phrasing, ordering of findings, and required components (lesion description, assessment, and recommendation). These are used to steer the LVLM during text generation.  

The two \emph{diagnostic} guidelines specify the clinical implications of relevant BI-RADS concepts (e.g., which findings suggest benign or malignant pathology, how specific descriptors alter risk). These are used by the LRM to produce guideline-aware diagnostic narratives.

\subsection*{B2. Sibley-Inspired Field Guide}
For the CUB-200 images, we provide a compact field-guide entry inspired by the Sibley bird identification manual. Each entry summarizes key visual traits (color, shape, pattern, behaviors) and their taxonomic relevance. These descriptions serve as lightweight analogues of clinical guidelines for non-medical domains. An example of the guideline can be seen in Figure~\ref{fig:birb_snippets}.

\begin{figure}[t!]
\centering
\small
\begin{lstlisting}[style=medcbr]
concept_data = "Finally, you are given the following 'concepts' " \
               "that are present in the image.\n"

for i in range(len(selected_concepts)):
    if metadata["concepts"][i] == 1:
        name = named_concepts[i].replace("_", " ").capitalize()
        concept_data += f"{name}: 1\n"

prompt = f"""
You are given the following {modality} <image>. {auxiliary_data}
You are also given the following {type_of_guideline} guideline:

{GUIDELINE}

{concept_data}

Write a report based on the image, the guideline provided, and the concepts present in the image.
"""

messages = [{
    "role": "user",
    "content": [
        {"type": "image", "image": image}, 
        {"type": "text", "text": prompt},
    ],
}]
\end{lstlisting}

\caption{LVLM prompting strategy used in our experiments. 
The LVLM receives the image, the appropriate reporting guideline 
(BI-RADS or field-guide), the predicted concepts, and a final 
instruction describing the reporting task.}
\label{fig:lvlm_prompt}
\end{figure}

\begin{figure}[t]
\centering
\small
\begin{lstlisting}[style=medcbr]
introduction = "You are given the final diagnostic prediction of an AI system, which is {diagnosis}. The system also detected the following concepts:\n"

concept_data = ""
for i, name in enumerate(named_concepts):
    score = metadata["concepts"][i] * 100
    if metadata["concepts"][i] >= 0.5:
        cname = name.replace("_"," ").capitalize()
        concept_data += f"{cname} ({score:.1f}% confidence)\n"
    else:
        if dataset_name == "BREAST_US" and name == "regular_shape":
            concept_data += f"Irregular shape ({score:.1f}% confidence)\n"

# for CUB, this would say "field guide"
instructions = "Assuming the diagnosis is correct, explain the implications of these concepts according to the BI-RADS clinical guideline provided. Interpret each concept, assess agreement with the predicted diagnosis, infer the most likely BI-RADS category, and provide a recommended follow-up.\n"

reasoning_prompt = f"""{introduction}
{concept_data}
{instructions}
{GUIDELINE}
"""
\end{lstlisting}

\caption{LRM prompting strategy. The LRM receives the model's final 
prediction, the predicted concepts, a domain-specific diagnostic or 
field-guide guideline, and instructions describing how to construct a 
reasoned explanation grounded in those concepts.}
\label{fig:lrm_prompt}
\end{figure}

\subsection*{B3. Prompting the LVLM}
\textbf{Prompt Structure.} For BUS-BRA and CBIS-DDSM, the LVLM is conditioned using the appropriate {reporting} guideline, which provides a structured template for BI-RADS–compliant report generation. For CUB-200, the LVLM instead receives the corresponding Sibley-inspired field-guide entry for the species, offering a compact attribute-oriented template. 

\textbf{Prompting Strategy.} We explored several prompting strategies to ensure that LVLM outputs were reliable and grounded.  
First, we supplied only the image and requested a clinical report; this produced text that was stylistically coherent but frequently inaccurate or logically inconsistent. Incorporating the BI-RADS reporting guideline improved the structure and style of the output but did not fully correct factual errors. We therefore augmented the prompt with the ground-truth concept labels to anchor the LVLM to verifiable findings, as shown in Figure~\ref{fig:lvlm_prompt}. For each configuration, outputs were evaluated by comparing the main assertions in the generated report with the clinical ground truth, and all reports were subsequently reviewed by a trained radiologist for fact-checking. 

\subsection*{B4. Prompting the LRM}
The LRM receives the predicted concepts from the concept-based model together with the appropriate \emph{diagnostic} guideline. These guidelines specify the clinical or taxonomic implications of each concept (e.g., “spiculated margins increase suspicion for malignancy” or “a blue crest is characteristic of a blue jay”), enabling the LRM to generate coherent reasoning narratives that reflect domain conventions. By grounding its explanation in both the structured predictions and the diagnostic rules, the LRM produces interpretable statements aligned with established clinical or field-guide knowledge. A snippet of our LRM prompting strategy can be seen in Figure~\ref{fig:lrm_prompt}.

\section*{C. Hyperparameter Tuning}

\textbf{Original CBM.}
For the baseline concept bottleneck model, we performed grid searches over batch size, learning rate, data augmentations, loss formulations, and optimizers. 
We evaluated two loss functions: 
(i) $\mu\mathcal{L}_{\text{CE}}(y,\hat{y}) + \nu\mathcal{L}_{\text{CE}}(c,\hat{c})$ and 
(ii) $\mathcal{L}_{\text{CE}}(y,\hat{y}) + \mathcal{L}_{\text{MSE}}(c,\hat{c})$. 
Formulation (i) performed consistently better, and we adopted $\mu=1.0$, $\nu=0.8$.  
Across optimizers (SGD, Adam, AdamW), AdamW yielded the most stable results.  
All models were trained for 150 epochs with early stopping based on validation loss.

\textbf{CLIP CBM.}
We evaluated several CLIP encoders ({ViT-B/32}, {ViT-L/14}, and {RN50}) as the visual backbone. CLIP {ViT-L/14} consistently achieved the strongest performance and was therefore used as the default.  We searched over the concept and label loss weights and retained $\mu = 1.0$ and $\nu = 1.0$.  All other hyperparameters followed the original CBM setup.

\begin{table*}[t]
\centering
\small
\begin{tabular}{p{4.5cm}|p{12.5cm}}
\toprule
\textbf{Metric} & \textbf{Description and Scoring Criteria} \\
\midrule

\textbf{Concept Interpretation Score (CIntS)} 
&
\textbf{Definition:} Measures whether each predicted concept is interpreted correctly according to BI-RADS semantics. \newline

\textbf{Score:} 
\[
\text{CIntS} = 
\frac{\text{\# correctly interpreted concepts}}
     {\text{\# predicted concepts}}
\]
\\
\midrule

\textbf{Concept Integration Score \newline (CIgS)} 
&
\textbf{Definition:} Evaluates whether multiple concepts are integrated coherently and consistently with BI-RADS guidelines. \newline

\textbf{Scoring Levels:} \newline
\begin{tabular}{p{1.2cm} p{10.6cm}}
\textbf{1.0 } & \textbf{Fully Correct:} All concept combinations follow BI-RADS guidelines; no contradictions. \\[0.2em]
\textbf{0.75 } & \textbf{Mostly Correct:} Most interpretations are reasonable, but some combinations deviate from BI-RADS (e.g., many malignant cues and a few misinterpreted concept combinations, yet the overall impression is still malignant). \\[0.2em]
\textbf{0.25 } & \textbf{Partially Correct:} Only a small portion is reasonable; several contradictions or guideline violations (e.g., several malignant cues and a few benign concept combinations, with a benign overall impression). \\[0.2em]
\textbf{0.0 } & \textbf{Incorrect:} Concept combinations or guideline references are entirely inconsistent or unreasonable. \\
\end{tabular}
\\
\midrule

\textbf{BI-RADS Assignment Score (BAS)} 
&
\textbf{Definition:} Measures whether the final BI-RADS category aligns with guideline-based decision criteria for the given case.\newline

\textbf{Scoring Levels:} \newline
\begin{tabular}{p{1.2cm} p{10.6cm}}
\textbf{1.0 } & \textbf{Correct Assignment:} Correct BI-RADS category chosen based on BI-RADS decision logic. \\[0.2em]
\textbf{0.8 } & \textbf{Near-Correct:} Unreasonable category, but correct malignant/benign implication (e.g., 4C vs.\ 5). \\[0.2em]
\textbf{0.0 } & \textbf{Incorrect:} Category contradicts BI-RADS criteria and results in an incorrect implication. \\
\end{tabular}
\\
\bottomrule
\end{tabular}
\caption{
Rubric used by the radiologist for evaluating the clinical validity of model reasoning. 
CIntS measures accuracy of individual concept interpretation, 
CIgS measures the coherence of multi-concept integration, 
and BAS evaluates correctness of the final BI-RADS assignment. 
}
\label{tab:reasoning_rubric}
\end{table*}

\textbf{Label-Free CBM.}
For LF-CBM, the primary hyperparameter is the strategy used to generate concept labels via GPT.  
Directly prompting GPT often resulted in redundant or noisy concepts, inflating the concept bank and reducing performance. 
We also tested bypassing GPT entirely and inserting curated concept labels before training using the authors' code.  
Performance varied by dataset, and we report the best-performing configuration in each case.

We additionally tuned the visual backbone and compared {CLIP RN50} and {CLIP ViT-L/14}.  
The optimal backbone varied by dataset; we selected the best-performing one for each experiment.

\textbf{AdaCBM.}
As with LF-CBM, the main hyperparameter in AdaCBM is the source of concept labels (GPT-generated vs.\ curated).  
On CUB-200, models using the GPT-derived concepts provided in the LF-CBM repository significantly outperformed those using the official ground-truth concept bank. We also tuned the visual backbone, comparing {CLIP ViT-B/32} and {CLIP ViT-L/14}. For each dataset, we retained the backbone that yielded the highest validation performance. 

AdaCBM uses a two-stage concept selection procedure: a t-test to filter non-informative concepts, followed by correlation-based redundancy removal.  
This process was often overly aggressive for CUB-200, removing nearly all concepts. 
We therefore tuned the \emph{interpretability cutoff} associated with the t-test threshold to reduce over-pruning.  
When the available concept bank was small, we adopted a very permissive (low) cutoff to preserve a sufficient number of concepts and improve downstream accuracy.

For each model and dataset, we report results from the best-performing configuration.

\begin{figure*}[h]
    \centering
    \includegraphics[width=\textwidth]{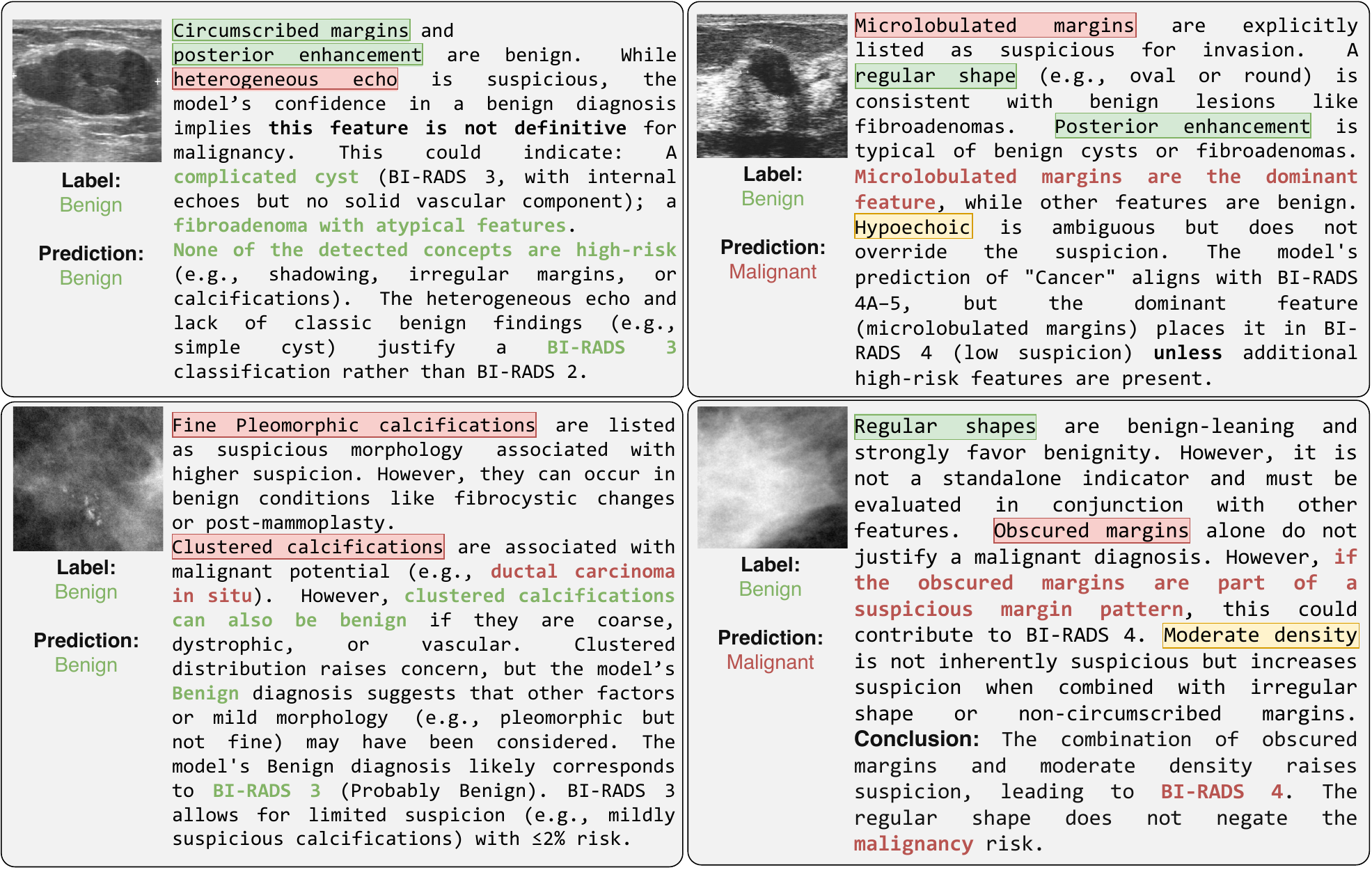}
    \caption{
    Additional examples of model reasoning on four challenging cases from BUS-BRA (top) and CBIS-DDSM (bottom). Malignant and benign-leaning clinical concepts are highlighted in \redbox{red} and \greenbox{green}, and their corresponding reasoning steps are emphasized. Neutral concepts are highlighted in \yellowbox{yellow}.
    }
    \label{fig:more_bca}
\end{figure*}
\begin{figure*}[t]
    \centering
    \includegraphics[width=\textwidth]{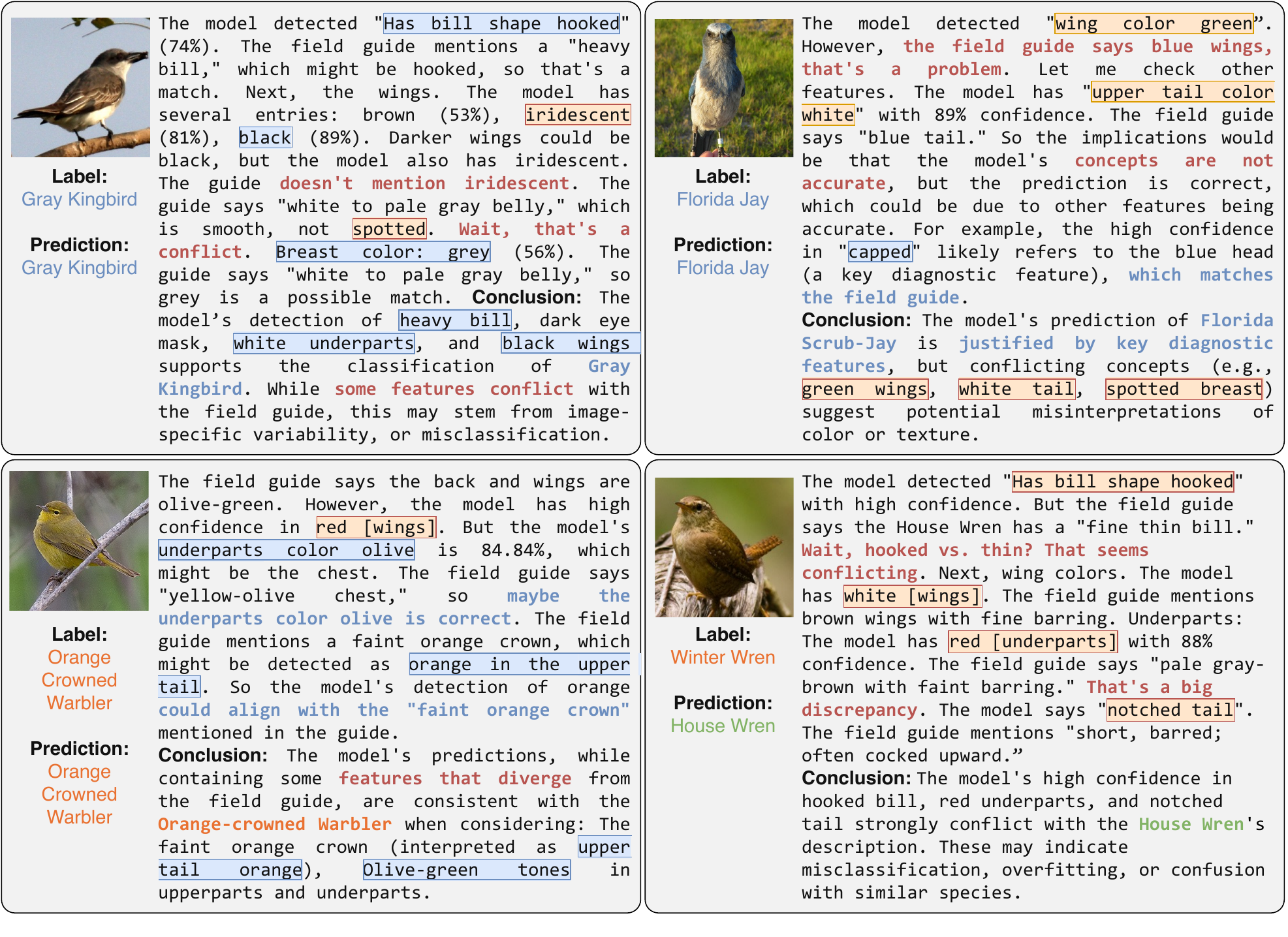}
    \caption{Additional examples of model reasoning on 4 complex cases from CUB-200. Observations highlighted in \orangebox{orange} contradict the guideline, whereas ones in \bluebox{blue} are consistent. Key reasoning steps are emphasized in \textbf{\textcolor{dioBlueStroke}{blue}} when they support the final conclusion and \textbf{\textcolor{dioRedStroke}{red}} when they do not.}
    \label{fig:more_birb}
\end{figure*}

\section*{D. Evaluation of Model Reasoning}

\subsection*{D1. Radiologist Evaluation}
To complement our quantitative reasoning metrics, we conducted an expert evaluation with an attending radiologist with specialized breast imaging training. The radiologist assessed 20 randomly selected cases drawn independently by a separate investigator to avoid bias.  For each case, the radiologist reviewed the predicted concepts, the LRM-generated reasoning statement, and the final diagnostic assignment, while remaining blinded to the ground-truth pathology and overall dataset composition. 

The assessment followed a structured evaluation rubric, which is outlined in full in Table~\ref{tab:reasoning_rubric}. The rubric evaluates (i) the \emph{Concept Interpretation Score (CIntS)}, measuring whether each concept is interpreted correctly within BI-RADS semantics; (ii) the \emph{Concept Integration Score (CIgS)}, which evaluates whether multiple concepts are combined logically and conflicts between benign and malignant cues are resolved; and (iii) the \emph{BI-RADS Assignment Score (BAS)}, which assesses whether the final diagnostic category is justified according to BI-RADS guideline criteria.

\subsection*{D2. Additional Examples of Model Reasoning}
We include additional qualitative reasoning examples on complex cases from BUS-BRA and CBIS-DDSM (shown in Figure~\ref{fig:more_bca}), as well as CUB-200 (shown in Figure~\ref{fig:more_birb}) to illustrate how MedCBR integrates concept predictions with domain-specific guidelines. Each example includes the image, instances of predicted concepts, the LRM-generated reasoning statement, and the final classification. 
For each of BUS-BRA and CBIS-DDSM, we highlight both a success case (left) and a failure case (right). For CUB-200, we highlight 4 complex cases where the concept model predictions are either partially correct or entirely incorrect, and show the LRM weighing conflicting evidence to support a coherent conclusion. \\
\indent\textbf{Reasoning on medical images.} In the top left image of Figure~\ref{fig:more_bca}, the model predicts a benign outcome and the LRM correctly explains that an isolated suspicious feature can still appear in benign lesions. In the top right image, the model incorrectly predicts malignancy; although the final diagnosis is benign, this case presents mixed or misleading features, leading to a higher risk score. In the bottom left image, despite multiple suspicious concepts, the model predicts benign correctly, and the LRM explains how these findings may still be compatible with a low-risk assessment. In the bottom right image, the model incorrectly predicts malignancy, and some detected concepts contradict the visual appearance (e.g., regular shape), leading the LRM to assign an overly high BI-RADS category.\\
\indent\textbf{Reasoning on natural images.} In Figure~\ref{fig:more_birb}'s top left example, the model detects the correct concepts, and generates a reliable explanation. In the subsequent two examples (top right and bottom left), the final prediction is correct, but the concepts contain multiple mistakes. These mistakes are then correctly identified by the LRM and highlighted to the user. Finally, in the bottom right case, the LRM rejects the concept model's prediction due to too many predicted concepts conflicting with the field guide.

\section*{E. Ethical Considerations}
First, there is a risk that clinicians may place undue trust in model-generated narratives, even when predictions are incorrect. Our framework is intended to support expert judgment, not replace it. Second, LLMs are computationally expensive to train and deploy, raising concerns about their environmental impact. Future work may explore efficient alternatives such as small language models (SLMs) or distillation-based techniques. Finally, while our method greatly reduces the likelihood of hallucinated outputs by conditioning on auxiliary information (e.g., predicted concepts), the risk of factual inaccuracy remains, especially in cases where the concept model fails to output reliable predictions. Additional safeguards and validation strategies are needed to ensure the reliability of generated narratives in clinical settings.

\end{document}